\documentclass[11pt]{article}
\usepackage[a4paper]{geometry}
\usepackage{clic2017}
\usepackage{times}
\usepackage{url}
\usepackage{latexsym}
\usepackage{graphicx}
\usepackage{xcolor}

\usepackage{array}
\newcolumntype{?}{!{\vrule width 2pt}}
\newcommand{\GG}[1]{}

\newcommand{\cut}[1]{}

\newcommand\blfootnote[1]{%
	\begingroup
	\renewcommand\thefootnote{}\footnote{#1}%
	\addtocounter{footnote}{-1}%
	\endgroup
}

\title{Overprotective Training Environments Fall Short at Testing Time:\\Let Models Contribute to Their Own Training}

\author{Alberto Testoni \\
  DISI, University of Trento \\
  {\tt alberto.testoni@unitn.it} \\\And
  Raffaella Bernardi \\
  CIMeC, DISI, University of Trento \\
  {\tt raffaella.bernardi@unitn.it} \\}

\date{}

\begin{document}
\maketitle
\global\csname @topnum\endcsname 0
\global\csname @botnum\endcsname 0
\begin{abstract}
  Despite important progress, conversational
  systems often generate dialogues that sound unnatural to humans. We
  conjecture that the reason lies in their different training
  and testing conditions: agents are  trained in
  a controlled ``lab'' setting but tested in the ``wild''. During
  training, they learn to generate an utterance given the
  human dialogue history.
On the other hand, during  testing, they must interact with each other, and hence deal with noisy data.
We
  propose to fill this gap by training the model with mixed batches containing both
  samples of human and machine-generated dialogues. We assess the
  validity of the proposed method on GuessWhat?!, a visual referential
  game.\blfootnote{Copyright ©2020 for this paper by its authors. Use permitted under Creative Commons License Attribution 4.0 International (CC BY 4.0).}

\end{abstract}

\section{Introduction}
\label{sec:introduction}

\begin{figure}[t]\centering 
	\includegraphics[width=1\linewidth]{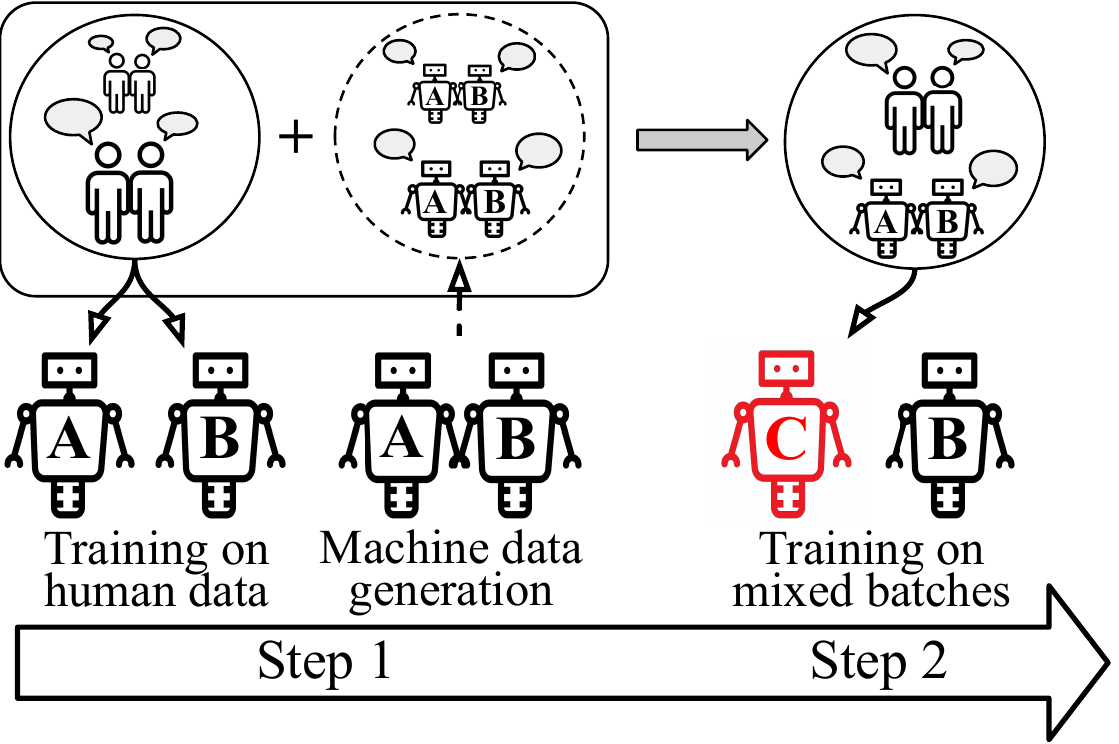}
	\caption{Two-steps training method of the C Bot: two Bots, A and B, are
       trained independently to reproduce human dialogues; then they play together to generate new dialogues (step
       1). In step 2 the Bot C is trained on mixed batches of human
       and machine-generated data (by A and
       B in step 1).}\label{fig:intro_abstract}
\vspace*{-5pt}
\end{figure}

\begin{figure*}[t]\centering 
	\includegraphics[width=1\linewidth]{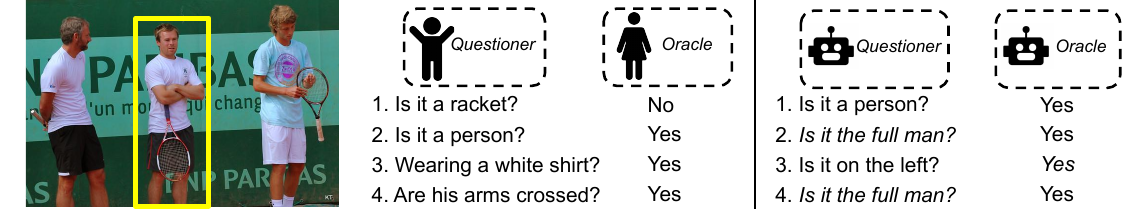}
	\caption{GuessWhat sample dialogues between two human annotators (left) and two conversational agents
          (right, generated by GDSE-SL as in Shekhar et al. \shortcite{shekhar-etal-2019-beyond}). The yellow box highlights the target entity that the Questioner has to guess by asking binary questions to the Oracle. Both humans and conversational agents have to guess the target object only at the end of the dialogue. Note that the machine-generated dialogue on the right contains repetitions from the Questioner and wrong answers from the Oracle (both in \textit{italic}). }\label{fig:intro}
\vspace*{-5pt}
\end{figure*}

Important progress has been made in the last years on developing conversational agents, thanks to the introduction of the encoder-decoder framework \cite{suts:seq14} that allows learning directly from raw data for both natural language understanding and generation. Promising results were obtained both for chit-chat \cite{Vinyals2015ANC} and task-oriented dialogues \cite{deal:lewi17}. The framework has been further extended to develop agents that can communicate about a visual content using natural language \cite{guesswhat_game,imag:mosta17,visdial}.
It is not easy to evaluate the performance of dialogue systems, but
one crucial aspect is the quality of the generated dialogue. These
systems must in fact produce a dialogue that sounds natural to humans
in order to be employed in real-world scenarios. Although there is not
a general agreement on what makes a machine-generated text sound
natural, some features can be easily identified: for instance, natural
language respects syntactic rules and semantic constraints, it is
coherent, it contains  words with different frequency
distribution but that crucially are informative for the conveyed message,
and it does not have repetitions, both at a token and a sentence level.

Unfortunately, even state-of-the-art dialogue systems often generate a
language that sounds unnatural to humans, in particular with respect
to the large number of repetitions contained in the generated
output. We conjecture that part of the problem is due to the training
paradigm adopted by most of the systems. In the Supervised Learning
training paradigm, the utterances generated by the models during
training are used only to compute a Log Likelihood loss function with
the gold-standard human dialogues and they are then thrown away. In a
multi-turn dialogue setting, for instance, the follow-up utterance is
always generated starting from the human dialogue and not from the
previously generated output. In this way, conversational agents never
really interact one with the other. This procedure resembles a
controlled ``laboratory setting", where the agents are always exposed
to ``clean" human data at training time. Crucially, when tested, the
agents are instead left alone ``in the wild", without any human
supervision. They have to ``survive" in a new environment
by exploiting the skills learned in the controlled lab setting and by
interacting one with the other. 


Agents trained in a Reinforcement Learning fashion are instead trained ``in the wild" by maximizing a reward function based on the task success of the agent, at the cost of a significant increase of computational complexity. Agents trained according to this paradigm generate many repetitions and the quality of the dialogue degrades. This issue is mildly solved by the Cooperative Learning training, but still, several repetitions occur in the dialogues, making them sound unnatural.

In this paper, we propose a simple but effective method to alter the training environment so that it becomes more similar to the testing one (see Figure \ref{fig:intro_abstract}). In particular, we propose to replace part of the human training data with dialogues generated by conversational agents talking to each other; these dialogues are ``noisy", since they may contain repetitions, a limited vocabulary etc. We then propose to train a new instance of the same conversational agent on this new training set. The model is now trained ``out of the lab" since the data it is exposed to are less controlled and they get the model used to live in an environment more similar to the one it will encounter during testing.

We assessed the validity of the proposed method on a referential
visual dialogue game, GuessWhat?!~\cite{guesswhat_game}. We found that the model trained according to our method outperforms the one trained only on human data with respect both to the accuracy in the guessing game and to the linguistic quality of the generated dialogues. In particular, the number of games with repeated questions drops significantly.

\section{Related Work}

\cut{Several dialogue tasks have been recently proposed as referential
  guessing games in which an agent (the Questioner) asks questions to
  another agent (the Answerer/Oracle) in order to guess the referent
  entity (e.g. a specific object depicted in an image) they have been
  speaking about
  \cite{visdial_rl,guesswhat_game,he:lear17,haber-etal-2019-photobook,ilinykh-etal-2019-tell,udag19}. These
  referential games are usually evaluated by looking at the task
  success metric, i.e. whether the Questioner is able to correctly
  identify the referent at the end of a dialogue exchanged with the
  other agent.  This simple task success metric, however, does not
  evaluate a core aspect of these systems, i.e. the linguistic quality
  of the generated output.}  The need of going beyond the task success
metric has been highlighted in Shekhar et
al.~\shortcite{shekhar-etal-2019-beyond}, where the authors compare
the quality of the dialogues generated by their model and other
state-of-the-art questioner models according to some linguistic
metrics. One striking feature of the dialogues generated by these
models is the large number of games containing repeated questions,
while the dialogues used to train the model (collected with human
annotators) do not contain repetitions. In Shekhar et
al.~\shortcite{shekhar2019jointly} the authors enrich the model
proposed in Shekhar et al.~\shortcite{shekhar-etal-2019-beyond} with a
module that decides when the agent has gathered enough
information and is ready to guess the target object. This approach is effective in reducing repetitions but,
crucially, the task accuracy of the game decreases. 

Murahari et
al.~\shortcite{mura:impr19} propose a Questioner model for the
GuessWhich task~\cite{visdial_rl} that specifically aims to improve
the diversity of generated dialogues by adding a new loss function
during training: the authors propose a simple auxiliary loss that penalizes similar dialogue state embeddings in consecutive turns. Although this technique reduces the number of
repeated questions compared to the baseline model, there is still a
large number of repetitions in the output. Compared to these methods, our method does not require to design ad-hoc loss functions or to plug
additional modules in the network.

The problem of generating repetitions not only affects dialogue
systems, but instead it seems to be a general property of current
decoding strategies. Holtzman et al. \shortcite{holt:cur19} found that
decoding strategies that optimize for an output with high probability,
such as the widely used beam/greedy search, lead to a linguistic
output that is incredibly degenerate. Although language models
generally assign high probabilities to well-formed text, the highest
scores for longer texts are often repetitive and incoherent. To
address this issue, the authors propose a new decoding strategy
(Nucleus Sampling) that shows promising results.

\section{Task and Models}

\paragraph{Task} The GuessWhat?! game (de Vries et al. 2017) is a cooperative two-player game based on a referential
communication task where two players collaborate to identify a
referent. This setting has been extensively used in human-human
collaborative dialogue \cite{clark1996using,yule2013referential}.  It is an asymmetric game involving two human
participants who see a real-world image. One of the participants (the Oracle) is secretly
assigned a target object within the image and the other participant
(the Questioner) has to guess it by asking binary (Yes/No) questions
to the Oracle.

\paragraph{Models} We use the Visually-Grounded State Encoder (GDSE)
model of Shekhar et al. \shortcite{shekhar-etal-2019-beyond}, i.e. a
Questioner agent for the GuessWhat?! game. We consider the version of
GDSE trained in a supervised learning fashion (GDSE-SL). The model
uses a visually grounded dialogue state that takes the visual features
of the input image and each question-answer pair in the dialogue
history to create a shared representation used both for generating a
follow-up question (QGen module) and guessing the target object
(Guesser module) in a multi-task learning scenario. More specifically, the visual features are extracted with a ResNet-152 network \cite{he2016:resnet}  and the dialogue history is encoded with an LSTM network. Since QGen faces a
harder task and thus requires more training iterations, the authors made the learning schedule task-dependent. They called this setup \textit{modulo-n} training, where \textit{n} specifies after how many epochs of QGen training the Guesser component is updated together with QGen. The QGen component
is optimized with the Log Likelihood of the training dialogues, and
the Guesser computes a score for each candidate object by performing
the dot product between visually grounded dialogue state and each
object representation. 
As standard practice, the dialogues generated
by the QGen are used only to compute the loss function, and the
Guesser is trained by receiving human dialogues. At test time, instead, the
model generates a fixed number of questions (5 in our work) and the
answers are obtained with the baseline Oracle agent presented in de Vries et
al. \shortcite{guesswhat_game}.  Please refer to Shekhar et
al. \shortcite{shekhar-etal-2019-beyond} for any additional detail on
the model architecture and the training paradigm.

\section{Metrics}
\label{sec:metrics}

The first metric we considered is the simple task accuracy (ACC) of the Questioner agent in guessing the target object among the candidates. 
We use four metrics to evaluate the quality of the generated dialogues. (1) Games with repeated questions (GRQ), which measures the percentage of games with at least one repeated question verbatim. (2) Mutual Overlap (MO), which represents the average of the BLEU-4 score obtained by comparing each question with the other questions within the same dialogue. (3) Novel questions (NQ), computed as the average number of questions in a generated dialogue that were not seen during training (compared via string matching). (4) Global Recall (GR), which measures the overall percentage of learnable words (i.e. words in the vocabulary) that the models recall (use) while generating new dialogues.
MO and NQ metrics are taken from Murahari et al.,
\shortcite{mura:impr19} while the GR metric is taken from van
Miltenburg et al., \shortcite{van-miltenburg-etal-2018-measuring}. We believe that, overall, these metrics represent a good proxy of the quality of the generated dialogues.

\section{Datasets}

\begin{table*}[]
	\centering
	\begin{tabular}{|c|c|c|c|c|c|}
		\hline
		\textbf{\begin{tabular}[c]{@{}c@{}}\% Human \\
				Dialogues\end{tabular}} &
		\textbf{\begin{tabular}[c]{@{}c@{}}\%
				Generated
				\\
				Dialogues\end{tabular}}
		&
		\textbf{\begin{tabular}[c]{@{}c@{}}Generated
				Dial. \\
				Length\end{tabular}}
		& \textbf{Voc size} &
		\textbf{MO}$\downarrow$
		& \textbf{GRQ}$\uparrow$ \\ \hline
		100
		& 0
		& variable
		& 10469 & 0.05 & 0              \\ \hline
		75
		& 25
		& fixed
		& 4642& 0.07& 2.9             \\ \hline
		75
		& 25
		& variable
		& 4646& 0.07& 2.6          \\ \hline
		50
		& 50
		& fixed
		& 4391& 0.08& 5.4              \\ \hline
		50
		& 50
		& variable
		& 4396 &0.07 & 4.7           \\ \hline
		0 & 100 &  fixed & 2586& 0.10&10.4\\\hline
		0 & 100 &  variable & 2680& 0.10&10.6  \\\hline
		
	\end{tabular}
	\label{tab:trainingset}
	\caption{Statistics of training sets built with different proportions of human machine-generated dialogues. Human data (100-0) vs. Mixed Batches (75-25, 50-50) vs. Fully
		Generated data (0-100). Voc size: size of the vocabulary used. GRQ: \% games with at least one repeated question verbatim. MO: Mutual Overlap. Refer to Section 4 for additional details on the metrics.}
\vspace*{-5pt}
\end{table*}


We are interested in studying how modifying part of the human data in
the training set affects the linguistic output and the model's
accuracy on the GuessWhat game. More specifically, we aim at building
a training set in which part of the dialogues collected with human
annotators are replaced with dialogues generated by the GDSE-SL
questioner model while playing with the baseline Oracle model on the
same games being replaced. In this way, we build a training set
containing dialogues that are more similar to the ones the model will
generate at test time while playing with the Oracle.

\paragraph{Human data}
The training set contains about 108K dialogues and the validation and
test sets 23K each. Dialogues contain on average 5.2 turns.  
The
GuessWhat?!\@ dataset was collected via Amazon Mechanical Turk by de
Vries et al. (2017). The images used in GuessWhat?!\@ are taken from the MS-COCO dataset
\cite{lin:micr14}. Each image contains at least three and at most twenty objects. More
than ten thousand people in total participated in the dataset collection procedure. Humans could stop asking questions at any time, so the
length of the dialogues is not fixed. Humans used a vocabulary of 17657 words  to play GuessWhat?!: 10469 of these words appear at least three times, and thus make up the vocabulary given to the models. For our experiments, we considered only those games in which humans succeeded in identifying the target object and that contain less than 20 turns.

\paragraph{Mixed Batches} We let the GDSE-SL model play with the baseline Oracle on the same
  games of the human training dataset. This produces
  automatically generated data for the whole training set. The model uses less than 3000 words out of a vocabulary of more than 10000 words. We built new training sets according to
two criteria: the proportion of human and machine-generated data
(50-50 or 75-25) and the length of the generated dialogue. Either we
always keep a fixed dialogue length (5 turns, as the average length in
the dataset) or we take the same number of turns that the human
Questioner used while playing the game we are replacing.

Table 1 reports some statics of different training sets. Human dialogues have a
very low mutual overlap and a much larger vocabulary than both the
generated (0-100) and mixed batches datasets (50-50, 75-25). Looking at the number of games
with at least one repeated question in the training set (GRQ column in Table 1), it can be observed that human
annotators never produce dialogues with repetitions. The 75/25 dataset
configuration contains less than 3\% of dialogues with repeated
questions and this percentage rises to around 5\% for the 50/50
configuration and to around 10\% for generated dialogues. Looking at the vocabulary size, the human dataset (100-0) contains around ten thousand unique words, the mixed batches datasets (50-50, 75-25) around 4500 words, and the generated dialogues (0-100) approximately 2500 words.

\begin{table*}[]
	\centering
	\begin{tabular}{|c|c|c|c|c|c|c|c|}
		\hline
		\textbf{\begin{tabular}[c]{@{}c@{}}\% Human \\ Dialogues\end{tabular}} & \textbf{\begin{tabular}[c]{@{}c@{}}\% Generated \\ Dialogues\end{tabular}} & \textbf{\begin{tabular}[c]{@{}c@{}}Generated Dial. \\ Length\end{tabular}} & \textbf{ACC$\uparrow$} & \textbf{GRQ$\downarrow$} & \textbf{MO$\downarrow$} & \textbf{NQ$\uparrow$} & \textbf{GR$\uparrow$} \\ \hline
		100                                                                    & 0                                                                          & variable                                                                                   & 46.3              & 36.8         & 0.27        & 0.53        & 20.6        \\ \hline
		75                                                                     & 25                                                                         & fixed                                                                         & 47.9              & 24.0           & 0.20        & 0.43        & 20.2        \\ \hline
		75                                                                     & 25                                                                         & variable                                                              & 47.5              & 26.6         & 0.21        & 0.41        & 19.4        \\ \hline
		50                                                                     & 50                                                                         & fixed                                                                       & 48.1              & 22.5         & 0.18        & 0.37        & 21.2        \\ \hline
		50                                                                     & 50                                                                         & variable                                                            & 47.0                & 21.0         & 0.18        & 0.42        & 21.1        \\ \hline
	\end{tabular}
	\label{tab_results}
	\caption{Test Set 5Q setting. GDSE-SL results on several training sets. At test time, the model generates 5 questions and then it guesses. Length ``fixed": 5-turns dialogues. Length ``variable": same turns human annotators used for that game. ACC: accuracy. GRQ: \% games with at least one repeated question. MO: Mutual Overlap. NQ: Novel Questions. GR: Global Recall. $\uparrow$: higher is better. $\downarrow$: lower is better.}
\vspace*{-5pt}
\end{table*}

\section{Experiment and Results}
\label{sec:results}

\subsection{Experiment}
As a first step, we trained the GDSE-SL model for 100 epochs as described in Shekhar et al. \shortcite{shekhar-etal-2019-beyond}. At the end of the training, we used GDSE to play the game with the Oracle on the whole training set, saving all the dialogues. We generate these dialogues with the model trained for all the 100 epochs since it generates fewer repetitions, although it is not the best-performing on the validation set. The dialogues generated by GDSE while playing with the Oracle are noisy: they may contain duplicated questions, wrong answers, etc. See Figure \ref{fig:intro} for an example of human and machine-generated dialogues for the same game. We design different training sets as described in Section 5 and train the GDSE-SL model on these datasets. We scrutinize the effect of training on different sets using the metrics described in Section 4 by letting the model generate new dialogues on the test set while playing with the Oracle. 

\subsection{Results}

Table 2 reports the results of the GDSE model trained on different training sets. To sum up, there are five dataset configurations: apart from the original GuessWhat dataset composed of dialogues produced by human annotators (100\% Human Dialogues), there are datasets composed of 75\% human dialogues and 25\% generated dialogues or 50\% human dialogues and 50\% generated dialogues. For each dataset configuration, the generated dialogues can be always 5-turns long (``fixed" length) or they can have the same number of turns human annotators used for that game (``variable" length). We do not report the results on the dataset composed of generated dialogues only since it leads to a huge drop in the accuracy of the guessing game.

By looking at the results on the test
set, we can see how even a small number of machine-generated dialogues affects the generation phase at test
time, when the model generates 5-turns dialogues and, at the end of
the game, it guesses the target object. First of all, it can be
noticed that the accuracy of GDSE-SL trained on the new datasets
outperforms the one trained on the original training set: in
particular, the accuracy of GDSE trained on 50\% human dialogues and
50\% 5-turns generated dialogues is almost 2\% higher (in absolute
terms) than the model trained only on human dialogues. The model seems
to benefit from being exposed to noisy data at training time to better
perform in the guessing game using the dialogues generated by the model itself while playing with the Oracle.

\cut{Looking at the number of games with at least one repeated
  question in the training set (GRQ column in Table 1-left), it can be
  observed that human annotators never produce dialogues with
  repetitions. The 75/25 dataset configuration contains less than 3\%
  of dialogues with repeated questions and this percentage rise to
  around 5\% for the 50/50 configuration. By looking at the results on
  the test set, we can see how the small number of dialogues with
  duplicated questions in the training set affects the generation
  phase at test time, where the model generates 5-turns dialogues and,
  at the end of the game, it guesses the target object. First of all,
  it can be noticed that the accuracy of GDSE trained on the new
  datasets outperforms the one trained on the original training set:
  in particular, the accuracy of GDSE trained on 50\% human dialogues
  and 50\% 5-turns generated dialogues is almost 2\% higher (in
  absolute terms) than the model trained only on human dialogues. The
  model seems to benefit from being exposed to noisy data at training
  time to better perform in the guessing game.}

The linguistic analysis of the dialogues generated on the test set reveals that the models trained on ``mixed" batches produce better dialogues according to the metrics described in Section \ref{sec:metrics}. In particular, considering the best-performing model on the test set, the percentage of games with repeated questions drops by 14.3\% in absolute terms and the mutual overlap score by 0.09. The percentage of vocabulary used (global recall), on the other hand, remains stable. Interestingly, the only metric that seems to suffer from the model being trained on mixed datasets is the number of novel questions in the generated dialogue: being trained on noisy data does not seem to improve the ``creativity" of the model, measured as the ability to generate new questions compared to ones seen at training time.

Overall, our results show an interesting phenomenon: replacing part of the GuessWhat?!\@ training set with machine-generated noisy dialogues, and training the GDSE-SL questioner model on this new dataset, is found to improve both the accuracy of the guessing game and the linguistic quality of the generated dialogues, in particular with respect to the reduced number of repetitions in the output.

\section{Conclusion}
\label{sec:conclusions}

Despite impressive progress on developing proficient conversational
agents, current state-of-the-art systems produce dialogues that do not
sound as natural as they should.
In particular, they contain a high number of repetitions. To address
this issue, methods presented so far in the literature implement new
loss functions, or modify the models'
architecture. When applied to referential guessing games, these
techniques have the drawback of gaining little improvement, degrading
the accuracy of the referential game, or producing incoherent
dialogues.  Our work presents a simple but effective method to improve
the linguistic output of conversational agents playing the GuessWhat?!\@
game. We modify the training set by replacing part of the dialogues
produced by human annotators with machine-generated dialogues. We
show that a state-of-the-art model benefits from being trained on this
new mixed dataset: being exposed to a small number of ``imperfect"
dialogues at training time improves the quality of the output
 without deteriorating its accuracy on the task.
Our results show an absolute improvement in
the accuracy of +1.8\% and a drop in the number of dialogues
containing duplicated questions of around -14\%. Further work is
required to check the effectiveness of this approach on other
tasks/datasets, and to explore other kinds of perturbations on the
input of generative neural dialogue systems.

\section*{Acknowledgements}

We kindly acknowledge the support of NVIDIA Corporation with the donation of the GPUs used in our research at the University of Trento. We acknowledge SAP for sponsoring the work.

\bibliography{main.bib}
\bibliographystyle{acl}

\end{document}